\documentclass{article}
\PassOptionsToPackage{square,numbers}{natbib}
\usepackage[preprint]{neurips_2025}
\usepackage[square,numbers]{natbib}
\usepackage[utf8]{inputenc} 
\usepackage[T1]{fontenc}    
\usepackage{hyperref}       
\usepackage{url}            
\usepackage{booktabs}       
\usepackage{amsfonts}       
\usepackage{nicefrac}       
\usepackage{microtype}      
\usepackage{xcolor}         
\usepackage{graphicx}
\usepackage{amsmath}
\usepackage{subcaption}
\usepackage{threeparttable}
\usepackage{multirow}
\usepackage{pifont}
\usepackage{wrapfig}
\usepackage{algorithm}
\usepackage{algorithmic}
\usepackage[toc,page]{appendix}

\newcommand{\cmark}{\ding{51}}%
\newcommand{\xmark}{\ding{55}}

    \title{TempRe: Template generation for single and direct multi-step retrosynthesis} 
     
    \author{Xuan Vu Nguyen \textsuperscript{1}, \quad
  Daniel Armstrong\textsuperscript{1},\quad Zlatko Jončev\textsuperscript{1},\quad Philippe Schwaller\textsuperscript{1,2} \\
  \textsuperscript{1}\'Ecole Polytechnique F\'{e}d\'{e}rale de Lausanne (EPFL) \\
  \textsuperscript{2}National Centre of Competence in Research (NCCR) Catalysis \\
  \texttt{\{nguyen.nguyen,philippe.schwaller\}@epfl.ch} \\
}

\begin{document}

\maketitle
\begin{abstract}
    Retrosynthesis planning remains a central challenge in molecular discovery due to the vast and complex chemical reaction space. While traditional template-based methods offer tractability, they suffer from poor scalability and limited generalization, and template-free generative approaches risk generating invalid reactions. In this work, we propose TempRe, a generative framework that reformulates template-based approaches as sequence generation, enabling scalable, flexible, and chemically plausible retrosynthesis. We evaluated TempRe across single-step and multi-step retrosynthesis tasks, demonstrating its superiority over both template classification and SMILES-based generation methods. On the PaRoutes multi-step benchmark, TempRe achieves strong top-k route accuracy. Furthermore, we extend TempRe to direct multi-step synthesis route generation, providing a lightweight and efficient alternative to conventional single-step and search-based approaches. These results highlight the potential of template generative modeling as a powerful paradigm in computer-aided synthesis planning.

\end{abstract}

\newpage
\section{Introduction}

Despite significant progress in organic synthesis \cite{colombo2008chemistry}, it remains a major bottleneck in the design-make-test-analyse (DMTA) cycle of molecular design \cite{nicolaou2014organic}.
In contrast to the well-established automation of peptide synthesis \cite{merrifield1965automated}, the synthesis of small-to-medium-sized organic molecules poses greater challenges. These challenges arise from the structural diversity of target compounds, the vast array of available building blocks, and the inherent complexity of chemical reactions. Moreover, the synthetic search space grows exponentially with the number of reaction steps.
To address this complexity, the task of synthesis planning has been formalised as \textit{retrosynthesis}—a process of iteratively deconstructing a target molecule into simpler precursors until accessible starting materials are reached \cite{corey1969computer,segler2018planning,coley2017computer}. 
Given the enormity of this search space, chemists have long envisioned the aid of computational tools to suggest the most prominent synthetic routes to be tried in laboratories \cite{corey1969computer,vleduts1963concerning}. 
This is the core concept of computer-aided synthesis planning (CASP), where the key component is a retrosynthesis model that identifies potential reaction sites and corresponding precursors for a given target molecule, either with or without reaction conditions \cite{coley2017computer,segler2018planning,schwaller2020predicting}.

To represent reactions to computers, chemists frequently use a combination of the Simplified Molecular Input Line Entry System (SMILES) \cite{weininger1988smiles} for reagents and products, and SMILES arbitrary target specification (SMARTS) \cite{daylight2019smarts} strings for reaction templates.
A template is a substructural pattern that describes a graph transformation of the reacting atoms and their neighbors, thus having fewer degrees of freedom compared with the full reaction.
Indeed, while the space of chemical reactions is essentially as large as the pseudo-infinite chemical space \cite{reymond2015chemical}, the space of reaction templates can be represented as a vast finite set of rules. 
In line with this notion, the selection of expert rules or reaction templates has been the cornerstone of retrosynthesis software since the dawn of the field \cite{szymkuc2016computer}.
However, one of the main disadvantages of template-based approaches is that they are restricted to a fixed template library, limiting scaling to larger reaction datasets.
With the advent of machine learning and deep learning, synthesis planning has increasingly been approached from the perspective of generative modelling, where precursors are predicted through a \emph{de-novo} generative process, as opposed to classification over a set of template-based rules\cite{lin2020automatic,liu2017retrosynthetic,zhong2022root,tu2022permutation}, offering flexibility at the risk of generating invalid or chemically implausible transformations.

\begin{figure}[t]
    \centering
    \includegraphics[width=0.95\textwidth]{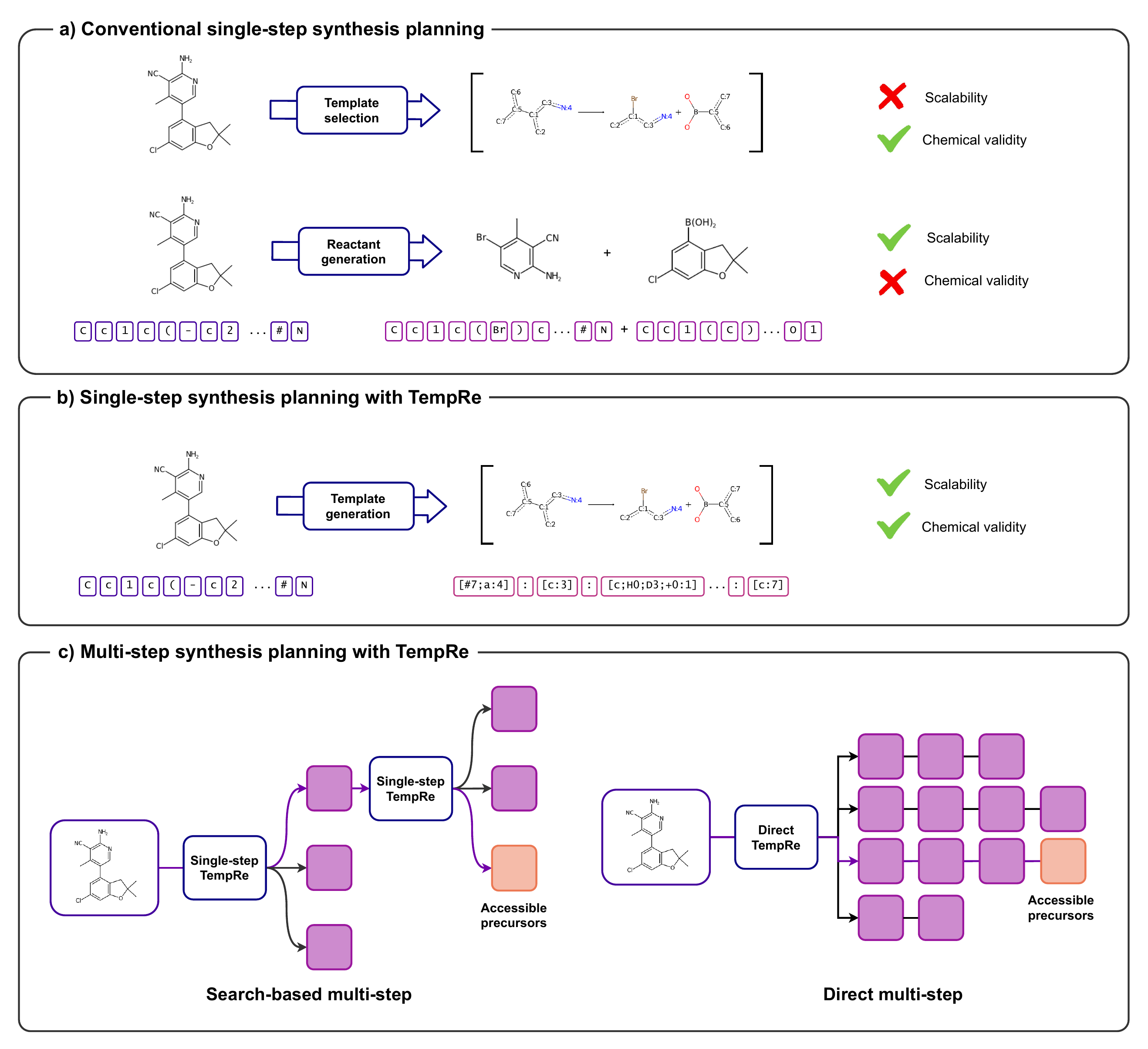}
    \caption{\textbf{Conceptual overview of TempRe}. a) Conventional single-step retrosynthesis paradigms based on template classification and template-free reactants generation. b) TempRe models a single-step transformation as translation-based template generation. c) TempRe as a framework for search-based and direct multi-step synthesis planning.}
    \label{fig:concept}
\end{figure}

Given the inherently linguistic nature of reaction templates \cite{thakkar2024neural}, this work studies a simple yet underexplored paradigm for retrosynthesis modeling: template generative retrosynthesis, or TempRe (Figure \ref{fig:concept}). 
We refer to TempRe not as a specific model, but as a general modeling practice that can be easily incorporated into existing sequence-based generative frameworks, such as Transformers \cite{vaswani2017attention, schwaller2019molecular, schwaller2020predicting}.
TempRe combines the advantages of both template-free and template-based approaches, while addressing their respective limitations. Like the former, TempRe can be trained on datasets containing an arbitrary number of reaction templates and is capable of generating novel templates not seen during training. At the same time, it retains the interpretability and structural validity checks inherent to template-based methods.

The key contributions of this work are as follows:
\begin{itemize}
    \item We perform a comprehensive evaluation of the TempRe paradigm, comparing it with traditional template classification and SMILES-based generation approaches across two primary application domains: single-step and multi-step retrosynthesis planning.
    \item We show that TempRe achieves strong performance on the common PaRoutes benchmark \cite{genheden2022paroutes}, even when trained on significantly noisier and more diverse datasets than those used in prior works, highlighting its robustness and scalability.
    \item We extend the TempRe framework to direct multi-step retrosynthesis by leveraging the straightforward sequential representation of a complete synthetic route as a series of reaction templates. To the best of our knowledge, this is the first work to apply template generation for direct multi-step synthesis planning.
\end{itemize}

\newpage
\section{Related Work}

\subsection*{Single-Step Retrosynthesis Models.}
Modern Computer-Assisted Synthesis Planning (CASP) is dominated by two main paradigms for single-step predictions. The classic approach is \textbf{template-based}, where chemical transformations are extracted from a reaction corpus and a model, typically a neural network, is trained to select the most appropriate template for a given target molecule \cite{segler2018planning, genheden2020aizynthfinder}. While robust and chemically interpretable, these methods struggle to generalise to rare or unseen transformations and scale poorly with the size of the template library. To address this, \textbf{template-free} methods frame retrosynthesis as a machine translation task, directly generating reactant SMILES strings from a product SMILES \cite{liu2017retrosynthetic, schwaller2019molecular}. The flexibility of this approach, especially with Transformer architectures, allows for the prediction of novel reactions but comes at the cost of often generating syntactically invalid or chemically implausible outputs \cite{liu2017retrosynthetic}. While advancements using graph-based representations \cite{tu2022permutation} or aligned SMILES \cite{zhong2022root} have sought to mitigate these issues, they often introduce significant model or preprocessing complexity.

\subsection*{Search-Based Multi-Step Synthesis Planning.}
A single-step model is typically embedded as a policy within a larger search algorithm to construct multi-step synthetic routes. Contemporary methods employ neural-guided graph exploration, most commonly Monte Carlo Tree Search (MCTS) \cite{segler2018planning} or A*-like algorithms on AND-OR graphs, such as Retro* \cite{chen2020retro}. These algorithms iteratively expand a search tree of intermediates until all branches terminate in commercially available starting materials. However, a recurring challenge in the field is that improvements in single-step prediction accuracy do not reliably translate to better multi-step planning performance \cite{torren2024models, li2024challenging, maziarz2025re}. While a conclusive explanation has yet to be established, it is not entirely surprising that models trained on isolated single-step reactions often fail to learn the multi-step strategic awareness required to solve complex synthesis planning tasks.

\subsection*{Generative Approaches to Templates and Routes.}
The inherent trade-off between the rigidity of template classification and the instability of template-free generation has motivated research into novel representations. One emerging area is \textbf{template generation}, where models autoregressively construct the reaction template (e.g., as a SMARTS string) instead of selecting it from a fixed list \cite{yan2022retrocomposer, shee2024site}. However, these pioneering works have only validated their approaches on single-step benchmarks, leaving their efficacy in multi-step planning unexplored. Concurrently, to overcome the limitations of iterative search, \textbf{direct multi-step} models aim to generate an entire synthetic route in a single inference step \cite{granqvist2025retrosynformer, shee2025directmultistep}. Current implementations often rely on complex and ambiguous route representations like nested JSON, which can be difficult for models to learn. This work addresses these gaps by proposing TempRe, a framework for generative template-based retrosynthesis. We provide the first comprehensive evaluation of template generation in a multi-step search context and extend the paradigm to direct multi-step planning by representing entire synthetic routes as a simple, powerful sequence of reaction templates.

\section{Methods}

\subsection{Template generative synthesis planning}
\label{sec:single_step_model}
For single-step synthesis planning, we denote target molecules as $o$ and reaction templates as $t$.
Since $t$ can be represented in a sequential format such as SMARTS, it can be tokenised into sequences of $N$ tokens $t=\{t_i\}_{i=1}^N$. Thus, the distribution of reaction templates $p_{\theta}(t|o)$, parameterised by $\theta$, can be modelled auto-regressively as:
\begin{equation}
\label{eq:autoregressive}
p_{\theta}(t|o)=\prod_{i=1}^N{p_{\theta}(t_i|t_{<i},o)}
\end{equation}
from which the distribution of reactant sets $R$ can be derived as:
\[ 
p(R|o)=\sum_{t}q(R|o,t)p_{\theta}(t|o)
\]
where $q(R|o,t)$ is a function that maps the reaction template $t$ and the target molecule $o$ to a set of reactants $R$. 
However, the combinatorial space of $\{t_i\}_{i=1}^N$ is intractable, even if we only consider syntactically valid SMARTS strings.
To ensure we only enumerate chemically valid templates, one could use a template library $T$ to limit the search space:
\[ 
p(R|o)=\sum_{t\in T}q(R|o,t)p_{\theta}(t|o)
\]
Unlike categorical classification models, the template library $T$ is an external component rather than a part of the model, thus the model size and architecture are independent of the template library.

If the target $o$ is also represented in sequential format (e.g., SMILES), Equation \ref{eq:autoregressive} can be modeled as a sequence-to-sequence Transformer, taking the target SMILES as input and outputting the reaction template SMARTS. We explored two primary tokenisation schemes for templates: Byte Pair Encoding (BPE) for the P2T model, and a frequency-sensitive tokeniser for P2T-Tok, which encodes frequent templates as single tokens and applies BPE to the remainder. Details are provided in Appendix \ref{app:tokenisation}. Input SMILES are tokenised as described in \cite{schwaller2019molecular}.
Post-inference, we introduced "strict" variants (P2T-Strict, P2T-Tok-Strict) that filter generated templates not present in the training set. Baselines include a product-to-reactant (P2R) sequence-to-sequence model and an AiZynthFinder (AZF) MLP template classifier, both trained on the same data. All sequence-to-sequence models utilise a Transformer architecture; hyperparameters and training specifics are in Appendix \ref{app:training_details}.
All TempRe models and baselines are summarised in Table \ref{tab:single-step_models}.

\begin{table}[h]
    \centering
    \caption{single-step model specifications}
    \label{tab:single-step_models}
    \resizebox{\columnwidth}{!}{%
    \begin{threeparttable}

    \begin{tabular}{@{}ll|llllll@{}}
        \toprule
        \multicolumn{2}{l|}{Model} & Input & Output & \begin{tabular}[c]{@{}l@{}}Template \\ tokeniser \tnote{a}\end{tabular} & \begin{tabular}[c]{@{}l@{}}Vocab \\ size\end{tabular} & \begin{tabular}[c]{@{}l@{}}Model\\ size\end{tabular} & \begin{tabular}[c]{@{}l@{}}Strict \\ filtering \tnote{b}\end{tabular} \\ \midrule
        \multirow{4}{*}{TempRe} & P2T & Product SMILES & Template SMARTS & BPE & 348 & 20M & \xmark \\
            & P2T-strict & Product SMILES & Template SMARTS & BPE & 348 & 20M & \cmark \\
            & P2T-Tok & Product SMILES & Template SMARTS & Frequency-based & 2651 & 22M & \xmark \\
            & P2T-Tok-strict & Product SMILES & Template SMARTS & Frequency-based & 2651 & 22M & \cmark \\ \midrule
        \multicolumn{2}{c|}{P2R} & Product SMILES & Reactant SMILES & N/A & 121 & 20M & \xmark \\
        \multicolumn{2}{c|}{AZF} & Product FPs & Template one-hot & N/A & 235K & 122M & \xmark \\ \bottomrule
    \end{tabular}%

    \begin{tablenotes}
        \small
        \item[a] BPE: Byte Pair Encoding; Frequency-based: Encode popular (>40 occurrences) reaction templates as one single token, and use BPE for the rest.
        \item[b] Filtering of post-inference templates not found in the training set.
    \end{tablenotes}

    \end{threeparttable}
    }
\end{table}

\subsection{Search-based multi-step synthesis planning}
For multi-step retrosynthesis, single-step models were integrated as policies within a Monte Carlo Tree Search (MCTS) algorithm, implemented using Syntheseus \cite{maziarz2025re}. Detailed parameters and implementation of the MCTS, including its phases, score functions, and search constants, can be found in Appendix \ref{app:search_details}.

\subsection{Direct multi-step synthesis planning}
We trained a variant of TempRe capable of predicting an entire synthetic route in a single inference, denoted as Direct TempRe. Instead of outputting a single retro-template, this model generates a sequence of reaction templates, which can be iteratively applied to the target molecule to construct the full route until starting materials are reached.


This model also employs the Transformer architecture with BPE tokenization for reaction templates, following a similar training procedure to the single-step TempRe models (details in Appendix \ref{app:training_details}). To mitigate the heavy bias towards short routes in the multi-step training data, we experimented with different inference strategies and additional input conditions, details of which are provided in Appendix \ref{app:direct_tempre_variants}.

To reconstruct synthetic routes from the generated template sequence, the templates are iteratively applied to the current molecular state, which is represented as a super node in a MolSet graph. This process, detailed in Algorithm \ref{alg:direct_tempre}, continues until a solved state (all molecules are in stock), an invalid template, or the end of the sequence is reached.

\subsection{Dataset}
\label{sec:data}
We used the USPTO dataset \cite{lowe2012extraction} for training and evaluation. Following a comprehensive preprocessing pipeline, we derived several data splits for different evaluation purposes, with full details on processing and benchmark construction provided in Appendix \ref{app:data_details}.

For single-step model evaluation, our primary test set was extracted from the PaRoutes benchmark \cite{genheden2022paroutes}. To further challenge the models, we created two additional test sets:
\begin{itemize}
    \item \textbf{A "hard" test set}, specifically constructed to be enriched with reactions that rely on rare templates, allowing us to assess model performance on less common chemical transformations.
    \item \textbf{An out-of-distribution (OOD) test set}, created by splitting data based on molecular weight to test generalization from smaller training molecules to significantly larger, unseen target molecules.
\end{itemize}

For training our Direct TempRe models, we curated a separate dataset of multi-step synthetic routes from the processed USPTO data. The detailed construction and characteristics of all datasets are available in Appendix \ref{app:data_details}.

\subsection{Baseline Models}
We compare our methods against established baselines for both single- and multi-step retrosynthesis.
For single-step planning, we use AiZynthFinder (AZF), a well-known template-based classifier, and a standard Product-to-Reactant (P2R) Transformer. The AZF baseline serves to highlight the parameter efficiency of our generative approach compared to classification methods that must handle a large template library.
For multi-step planning, we benchmark against popular search algorithms, Retro* and MCTS, which both use the AZF model for reaction proposals. For direct synthesis planning, we compare against the Direct Multi-Step (DMS) model \cite{shee2025directmultistep}. Detailed specifications and citations for all baseline models are provided in Appendix \ref{app:baseline_details}. 
We refrain from a wider analysis of single-step retrosynthesis models as in this work we aim to exclusively focus on the effect of decoding method (P2T/P2R) on single and multistep retrosynthesis performance.

\subsection{Evaluation}
\paragraph{Single-step Retrosynthesis}
Single-step models are evaluated by top-k accuracy \cite{schwaller2022machine} on the PaRoutes single-step test set and the hard test set. This metric measures the recall of ground-truth precursors among the top-k predictions. Invalid SMILES/SMARTS are filtered and duplicates removed. For template-based models (TempRe and AZF), predicted templates are applied to target molecules using RDChiral \cite{coley2019rdchiral} to obtain precursor sets. Top-k accuracy is calculated at the precursor level using a pessimistic setting \cite{roh2025higher} where ground-truth precursors from reactions applicable at multiple sites are ranked last.

\paragraph{Multi-step Retrosynthesis}
Evaluation of multi-step synthesis planning is conducted using the PaRoutes N1 and N5 benchmarks, with performance measured by solve-rate and top-k route accuracy. The N1 benchmark consists of 10,000 synthesis routes, each derived from a distinct US patent. The N5 benchmark, however, allows for up to five routes from the same patent, resulting in a dataset characterised by pathways of greater length and a higher incidence of 'convergent' structures. A target molecule is "solved" if the search algorithm finds a synthetic route where all precursors are in the provided stock set. Solve-rate is the percentage of solved targets. The stock sets for n1 and n5 contain 13432 and 13326 molecules, respectively.

Top-k route accuracy measures how well models retrieve the ground-truth synthetic routes among the top-k predictions. Route similarity is determined by Tree Edit Distance (TED) \cite{genheden2022fast,genheden2021clustering}, where a distance of zero signifies an exact match.

For search-based multi-step algorithms, predicted routes are ranked using a recursive scoring method from the PaRoutes package \cite{badowski2019selection}. The cost for an intermediate molecule $m$ is defined as:
\[
    \text{cost}(m) = \epsilon(r) + \sum_{m' \in \text{children}(m)} \frac{\text{cost}(m')}{\text{yield}(m')}
\]
where $\epsilon(r)$ is the cost of performing reaction $r$, and $\text{yield}(m')$ is the yield of precursor $m'$. Default parameters are $\epsilon(r)=1$ and $\text{yield}(m')=0.8$.

For Direct TempRe models, which auto-regressively output reaction templates for full routes, we use a ranking scheme based on route likelihood. Routes are ranked by prioritizing solved routes, then shorter routes, and finally routes with higher log-likelihood:
\[
    \text{rank}(R) < \text{rank}(R') \Leftrightarrow \begin{cases}
        \text{solved}(R) > \text{solved}(R'), & \text{if } \text{solved}(R) \neq \text{solved}(R') \\
        \text{len}(R) < \text{len}(R'), & \text{if } \text{solved}(R) = \text{solved}(R') \text{ and } \text{len}(R) \neq \text{len}(R') \\
        \text{log }p(R|o) > \text{log }p(R'|o), & \text{otherwise}
    \end{cases}
\]
where $\text{solved}(R)$ indicates if all leaf molecules are in-stock, $\text{len}(R)$ is the number of steps, and $\text{log }p(R|o)$ is the route log-likelihood.

\section{Experiments}

We aim to assess the performance of TempRe in four areas
\begin{itemize}
    \item How do various methods of template prediction and generation perform compared to traditional sequence-to-sequence models in the single-step retrosynthesis task?
    \item Sequence-to-sequence models have been shown to underperform in single-step retrosynthesis when the input molecules have high molecular weight \cite{gil2023holistic,chen2024assessing}. As the string length of the output of template generation is not correlated with the length of the input SMILES, does template generation outperform SMILES generation for high molecular weight inputs? Additionally, how well does template generation perform for low-frequency templates?
    \item How do methods of template prediction adapt to the search algorithm based multi-step retrosynthesis setting?
    \item Is it possible to directly generate entire synthesis routes as sequences of templates in an auto-regressive fashion, and how does this method's performance compare with search algorithm based approaches?
\end{itemize}

\section{Results \& Discussion}

\subsection{Single-step retrosynthesis}

\begin{figure}[t]
    \centering
    \includegraphics[width=\textwidth]{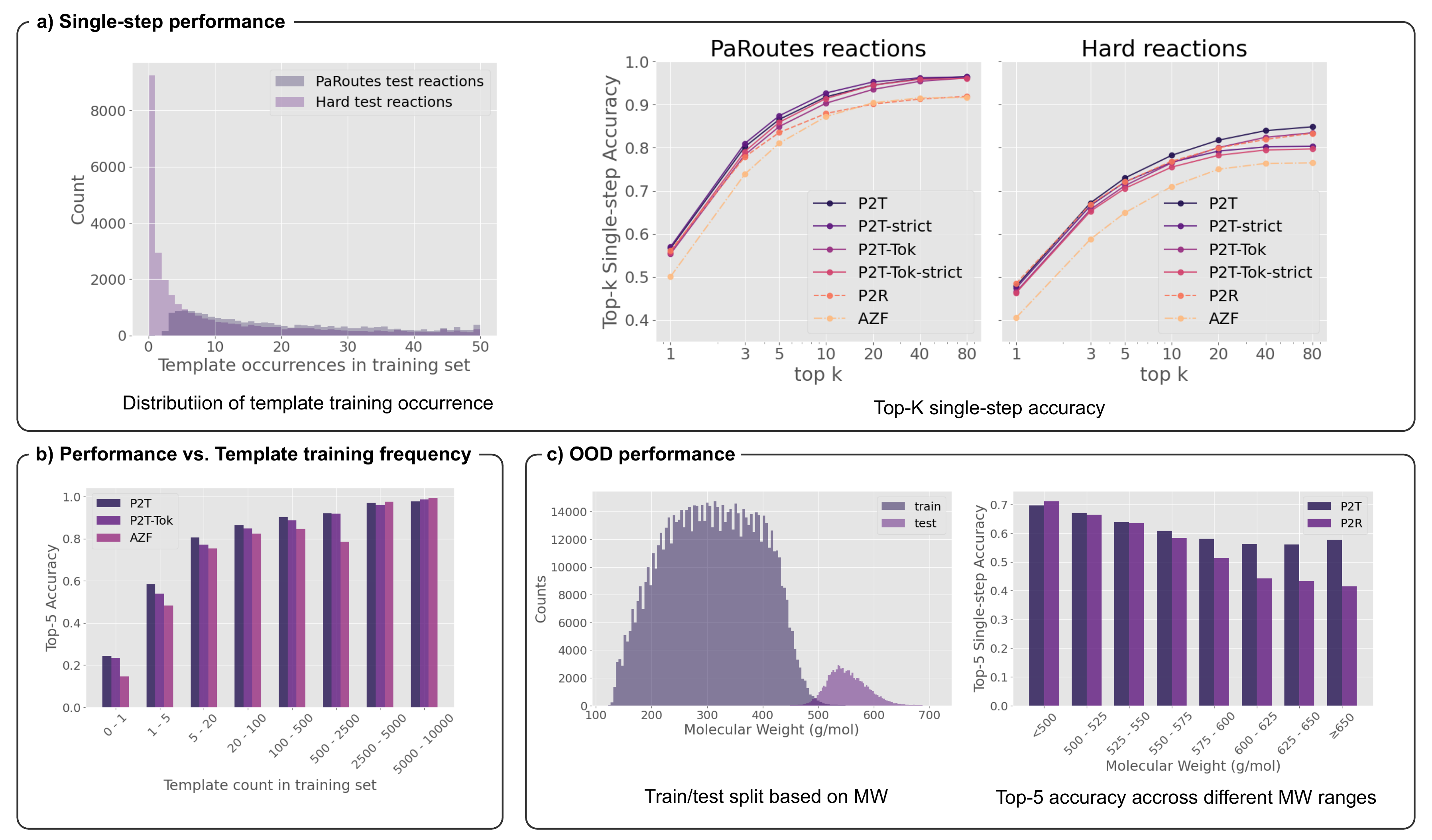}
    \caption{
        \textbf{Single-step synthesis planning performance}. a) Distribution of training occurrences of reaction templates of the PaRoutes test reactions and the "hard" test reactions, and the corresponding top-k reaction accuracy. The latter contains more rare or unseen reactions during training. b) Single-step model performance as a function of template training frequency. c) Top-5 accuracy of P2T and P2R in the OOD setting.
    }
    \label{fig:single_step}
\end{figure}

We compared the top-k accuracy of trained models on two test sets: 55K single-step reactions from PaRoutes sets N1 and N5, and 52K challenging reactions from the hard test set (Figures \ref{fig:single_step}a, b). On PaRoutes test reactions, AZF exhibited poor accuracy at lower values of top-k compared to other generative models, although performance converges with P2R from Top-10 onwards, with both models achieving approximately 92\% at Top-80. TempRe models across the board outperform SMILES-based sequence-to-sequence (P2R) models, with the gap widening after Top-5, reaching 96\% accuracy at Top-80. We note that although the difference is small, the Strict models perform better.

As template classification relies on previously seen templates to recovery the ground truth reactions, we expect generative approaches to exhibit stronger performance than template classification approaches. On the challenging test set, which contains reactions with rare or unseen reaction templates, performance declined across all models compared to PaRoutes. However the unrestricted generative models (P2T, P2T-Tok, P2R) achieve superior accuracy due to their greater ability to handle rare transformations. P2T achieved 85\% top-80 accuracy, followed by P2T-Tok and P2R at 83\%. The constrained models performed relatively worse, with P2T-strict and P2T-Tok-Strict providing 80\% accuracy and AZF achieving 76\%

We additionally perform analysis on model performance across different template frequency buckets. In buckets where templates appeared more than 2500 times in training data, all models performed comparably, whilst TempRe models substantially outperformed AZF for rare templates (Figure \ref{fig:single_step}b). This advantage likely stems from template selection models treating reaction templates as independent classes, preventing exploitation of semantic correlations between templates that SMARTS-trained models can leverage.



A well-known limitation of sequence-to-sequence models, including chemical language models, is poor generalisation to output sequences longer than those seen during training \citep{gil2023holistic,chen2024assessing}. This "length generalisation" problem poses a significant challenge for retrosynthesis, where P2R models must generate the SMILES strings of reactants using the product information. As target molecules become larger with more structural complexity, their corresponding reactants also tend to be larger, requiring the model to generate longer 
output sequences.
We propose that a P2T approach, where templates are generated using product information, can effectively bypass this limitation. Unlike P2R, the P2T model's task is to predict a reaction template. As the reaction template focuses exclusively on the reaction centre, the length and complexity of the output sequence is decoupled from the input SMILES length.  We hypothesise that a P2T model will exhibit superior robustness on large, out-of-distribution (OOD) molecules compared to a P2R model.
To test this hypothesis, we designed an OOD experiment. Figure \ref{fig:single_step}c clearly shows the separation in molecular weight (MW) distribution between our training set (mostly < 500 g/mol ) and our test set (entirely > 500 g/mol ), ensuring we evaluate generalization to larger molecules.
The presented results strongly support our hypothesis. While both models perform comparably on molecules just outside the training distribution (<500 g/mol, ~70\% Top-5 accuracy), their performance diverges dramatically as MW increases. The P2R model's accuracy steadily deteriorates, falling to just 42\% for molecules >=650 g/mol. In contrast, the P2T model’s performance remains relatively stable, maintaining an accuracy of ~57\% even in the highest MW bins.
This demonstrates that by reframing the retrosynthesis task to predict a size-agnostic reaction template, the P2T model successfully mitigates the length generalisation problem. This robustness is important for real-world applications, where target molecules can be substantially larger than those in conventional training sets.

\subsection{Search-based multi-step retrosynthesis}

\begin{figure}[t]
    \centering
    \includegraphics[width=0.9\textwidth]{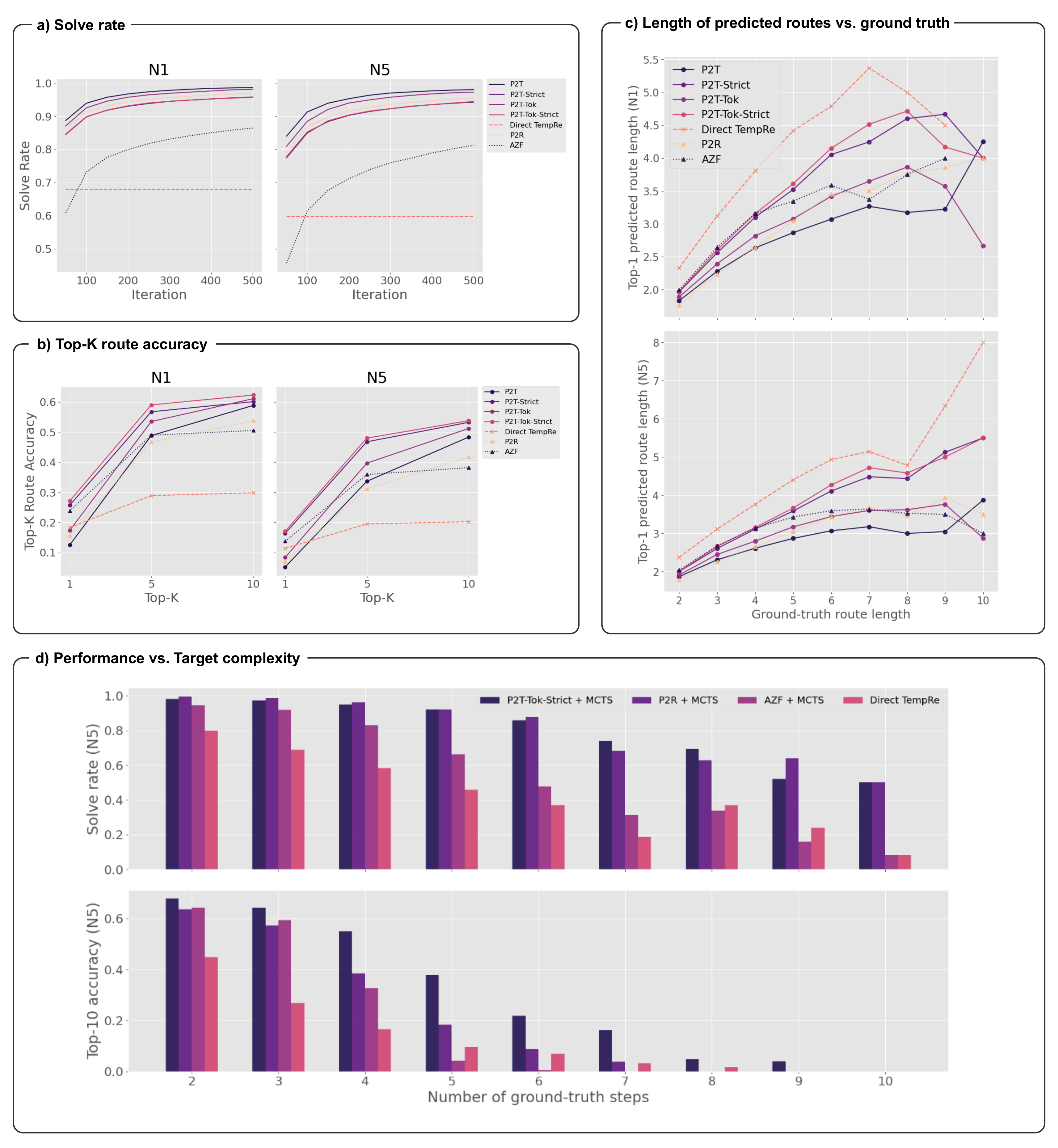}
    \caption{\textbf{Multi-step synthesis planning performance on PaRoutes benchmark}. a) Solve rate on n1 and n5 sets. b) Top-k route accuracy on n1 and n5 sets. c) Average number of steps in the top-ranked predicted route in comparison. d) Solve rate and top-k route accuracy in correlation with the number of reaction steps of the n5's ground-truth routes.}
    \label{fig:multistep}
\end{figure}

Having established the robustness of the P2T framework in single-step predictions, we sought to evaluate its efficacy as a policy model within a multi-step retrosynthesis search. We assessed performance on the PaRoutes benchmark, where finding a complete, viable route is a considerably more complex task.
Our initial evaluation focused on solve rate, the ability to find any valid synthetic route. The P2T-based models proved highly effective, with all variants achieving solve rates above 90\% and comfortably outperforming the AZF baseline (Figures \ref{fig:multistep}a). It is noteworthy that even when restricted to the same set of known templates as AZF, the P2T-Strict models achieved a substantially higher solve rate, indicating a more effective policy.
However, solve rate alone can be a misleading metric, as it may be inflated by the proposal of chemically plausible but practically suboptimal reactions that nonetheless lead to purchasable precursors. A more nuanced perspective is provided by top-k route accuracy, which measures the ability to identify the ground-truth synthesis. Here, we observed an inversion of performance (Figures \ref{fig:multistep}b). AZF, despite its low solve rate, yielded higher Top-1 accuracy than P2R and non-strict TempRe models. Yet, its performance plateaued, and by Top-10, all P2T variants surpassed it.
Crucially, the template-restricted models (P2T-Strict, P2T-Tok-Strict) consistently outperformed their unrestricted counterparts in route accuracy. This suggests a fundamental trade-off: the generative freedom of unrestricted models increases route diversity and thus the probability of finding a solution (high solve rate), but this appears to come at the cost of converging upon the ground-truth solution. The superior performance of P2T-Tok over P2T further supports this, as its single-token encoding likely biases generation towards more conventional, training-like templates.
To further scrutinise route quality, we examined the length of the top-ranked predicted path (Figure \ref{fig:multistep}c). 
Models restricted to known templates (AZF, P2T-Strict) produced routes with step counts closer to the ground truth. In contrast, unrestricted models often predicted shorter, more direct routes. The tendency for multiple models to find routes shorter than the recorded ground truth may also call into question the use of route length as a simple proxy for quality, as it overlooks real-world constraints such as yield and reagent availability that inform the curated solution.



\subsection{Direct multi-step retrosynthesis}

In addition to search-based methods to find synthetic routes, recent research has introduced direct multi-step approaches, which aim to generate an entire synthetic route in a single model call, eliminating the need for CPU-intensive search algorithm iterations \citep{granqvist2025retrosynformer, shee2025directmultistep}. To train such models, we constructed multi-step synthetic routes by chaining single-step reactions derived from our post-processed PaRoutes dataset. This process is done by linking the products of reactions to the reactants of others, and then performing a graph traversal to extract the longest route. A preliminary analysis reveals a strong bias in the training data towards short, two-step routes (Figure \ref{fig:data_distribution}b), which caused a vanilla model to perform poorly (Figure \ref{fig:direct_tempre_ablation}).

We explored several training strategies to address the short-route bias in our training data and tested two different data processing approaches (see Appendix~\ref{app:direct_tempre_variants} for full details). The most effective approach involved conditioning the model on the number of steps and scanning across different step counts (2-9) during inference. We refer to this as the "N-step variant" and use it as our representative direct synthesis planning method.

We trained this N-step model on two datasets with different processing strategies: our stricter data processing (Direct TempRe) and the same data processing used by \citet{shee2025directmultistep} (Direct TempRe-DMS). The latter achieved higher performance (0.75 solve rate vs 0.68, 0.45 top-20 accuracy vs 0.30) (Figure \ref{fig:direct_tempre_ablation}a). The solve rate boost on DMS data is likely due to data leakage - while we remove all single-step reactions found in the PaRoutes' testsets from the training routes, DMS only splits at the route level, which may allow leakage at the reaction level.

A full performance summary for all multi-step models is presented in Tables \ref{tab:multi-step_n1} and \ref{tab:multi-step_n5} for the n1 and n5 test sets, respectively. Our analysis reveals two key findings regarding the capabilities of direct and search-based approaches.

First, within the category of direct models, our template-based representation demonstrates a clear advantage. When trained on the same DMS dataset to ensure a fair comparison, Direct TempRe-DMS consistently outperforms the first direct route generation method, DMS's Explorer \citep{shee2025directmultistep}. On the n1 set, it achieves a 0.75 solve rate and 0.45 top-10 accuracy, surpassing Explorer's 0.74 and 0.40. This trend holds on the n5 set, underscoring the effectiveness of reaction templates over the nested JSON format for direct retrosynthesis.

Second, despite this success, a substantial performance gap remains between even the best direct models and leading search-based methods. This is most evident when comparing our Direct TempRe model trained on our stricter, non-leaky dataset with MCTS (P2T-Tok-Strict). On the N1 set, the search-based model achieves a 0.96 solve rate and 0.62 top-10 accuracy, significantly eclipsing Direct TempRe's 0.68 solve rate and 0.30 top-10 accuracy. We attribute this disparity to two fundamental architectural advantages of search-based models: they leverage a stock set to provide a reward signal that guides the search, and their underlying single-step models are trained on a larger corpus of reaction data.

Even with these inherent limitations, Direct TempRe proves to be a robust model, successfully providing solutions for 0.68 and 0.60 of targets in n1 and n5, respectively, establishing a strong performance baseline for direct models trained on rigorously cleaned data.
Additionally, by directly maximizing the likelihood of the whole synthetic routes, Direct TempRe's predicted routes are more similar to ground-truths in terms of route length compared with search-based methods, whose predicted routes are biased by the search algorithm (Figure \ref{fig:multistep}c).

The MCTS with P2T-Tok-Strict model establishes a new baseline for search-based methods on both the PaRoutes-n1 and PaRoutes-n5 benchmarks. On the n1 set, it achieves top-1, top-5, and top-10 accuracies of 0.27, 0.59, and 0.62, respectively, surpassing the implementation of AZF trained on the less noisy single-step reaction data of PaRoutes \cite{genheden2022paroutes}.
The model's superior performance is also evident on the n5 set, where it obtains top-k accuracies of 0.17 (top-1), 0.48 (top-5), and 0.54 (top-10).

\begin{table}[h]
    \centering
    \caption{Solve rate and top-k accuracy on PaRoutes' test set-n1. The best metric in each model category, search-based and direct, are highlighted in bold.} 
    \label{tab:multi-step_n1}
    \resizebox{\columnwidth}{!}{%
    \begin{threeparttable}
    \begin{tabular}{llcccccc}
    \toprule
    Model & \begin{tabular}[c]{@{}c@{}}Training\\ data\end{tabular}& \# templates & \begin{tabular}[c]{@{}c@{}}Size\\ (parameters)\end{tabular} & \begin{tabular}[c]{@{}c@{}}Solve \\ rate\end{tabular} & Top-1 & Top-5 & Top-10 \\ 
    \midrule
    MCTS (P2T-Tok-Strict) \textit{(SB)} & TempRe & 235K & 22M & 0.96 & \textbf{0.27} & \textbf{0.59} & \textbf{0.62} \\ 
    MCTS (P2R) \textit{(SB)} & TempRe & 235K & 20M & \textbf{0.97} & 0.16 & 0.47 & 0.54 \\ 
    MCTS (AZF) \textit{(SB)} & TempRe & 235K & 122M & 0.86 & 0.24 & 0.49 & 0.51 \\
    MCTS (AZF) \textit{(SB)} \tnote{a} & PaRoutes & 43K & 22M & \textbf{0.97} & 0.24 & 0.51 & 0.54 \\
    \midrule
    Direct TempRe \textit{(D)} & TempRe & - & 20M & 0.68 & 0.18 & 0.30 & 0.30 \\
    Direct TempRe-DMS \textit{(D)} & DMS & - & 20M & \textbf{0.75} & \textbf{0.29} & \textbf{0.44} & \textbf{0.45} \\
    DMS's Explorer \textit{(D)} \tnote{b} & DMS & - & 19M & 0.74 & 0.27 & 0.39 & 0.40 \\ 
    \bottomrule
    \end{tabular}%
    \begin{tablenotes}
        \small
        \item SB = Search-based; D = Direct multi-step
        \item[a] Data are collected from GitHub repository of PaRoutes version 2.0 \cite{genheden2022paroutes}
        \item[b] Data are produced by \citet{shee2025directmultistep}
    \end{tablenotes}
    \end{threeparttable}
       }
\end{table}

\begin{table}[h]
    \centering
    \caption{Solve rate and top-k accuracy on PaRoutes' test set-n5. The best metric in each model category, search-based and direct, are highlighted in bold.}
    \label{tab:multi-step_n5}
    \resizebox{\columnwidth}{!}{%
    \begin{threeparttable}
    \begin{tabular}{llcccccc}
    \toprule
    Model & \begin{tabular}[c]{@{}c@{}}Training\\ data\end{tabular}& \# templates & \begin{tabular}[c]{@{}c@{}}Size\\ (parameters)\end{tabular} & \begin{tabular}[c]{@{}c@{}}Solve \\ rate\end{tabular} & Top-1 & Top-5 & Top-10 \\  
    \midrule
    MCTS (P2T-Tok-Strict) \textit{(SB)} & TempRe & 235K & 22M & 0.94 & \textbf{0.17} & \textbf{0.48} & \textbf{0.54} \\ 
    MCTS (P2R) \textit{(SB)} & TempRe & 235K & 20M & 0.95 & 0.07 & 0.31 & 0.42 \\ 
    MCTS (AZF) \textit{(SB)} & TempRe & 235K & 122M & 0.81 & 0.14 & 0.36 & 0.38 \\
    MCTS (AZF) \textit{(SB)} \tnote{a} & PaRoutes & 43K & 22M & \textbf{0.97} & 0.12 & 0.36 & 0.41 \\ 
    \midrule
    Direct TempRe \textit{(D)} & TempRe & - & 20M & 0.60 & 0.11 & 0.19 & 0.20 \\
    Direct TempRe-DMS \textit{(D)} & DMS & - & 20M & \textbf{0.73} & \textbf{0.26} & \textbf{0.42} & \textbf{0.42} \\
    DMS's Explorer \textit{(D)} \tnote{b} & DMS & - & 19M & 0.72 & 0.25 & 0.36 & 0.37 \\ 
    \bottomrule
    \end{tabular}%
    \begin{tablenotes}
        \small
        \item SB = Search-based; D = Direct
        \item[a] Data are collected from GitHub repository of PaRoutes version 2.0 \cite{genheden2022paroutes}
        \item[b] Data are produced by \citet{shee2025directmultistep}
    \end{tablenotes}
    \end{threeparttable}
    }
    \end{table}


To assess model performance on increasingly complex targets, we analyse the solve-rate as a function of the ground-truth route length for the N5 test set (Figure \ref{fig:multistep}d). Here, the ground-truth route length serves as a proxy for synthetic complexity.
Unsurprisingly, the solve-rate for all models declines as route length increases. However, the P2T-Tok-Strict and P2R models exhibit a more graceful degradation in performance compared to AZF and Direct TempRe. For targets with 7 ground-truth steps, the solve-rates of AZF and Direct TempRe reduce to around 30\% and 20\%, respectively, while P2T-Tok-Strict and P2R maintain a solve-rate close to 80\%. 

This trend of performance degradation is also reflected in the top-10 route accuracy, which declines for all models on longer routes.
However, the top-10 accuracy metric reveals a divergence between P2T-Tok-Strict and P2R, exposing the limitations of using solve-rate as the sole measure of performance - P2R fails to find the correct route for any target with more than 7 steps. This result strongly suggests that its high solve-rate is a misleading metric, likely inflated by its ability to find any valid route rather than the chemically optimal or ground-truth one.
Among the four models, P2T-Tok-Strict exhibits the slowest decay rate in top-10 accuracy.
Moreover, P2T-Tok-Strict and Direct TempRe are the only two models with non-zero accuracy when solving routes with more than 7 steps. Collectively, these findings underscore the superior robustness of the TempRe framework, particularly the P2T-Tok-Strict variant, for modeling long and complex synthetic routes where other approaches falter.

\begin{figure}[h!]
    \centering
    \includegraphics[width=0.9\textwidth]{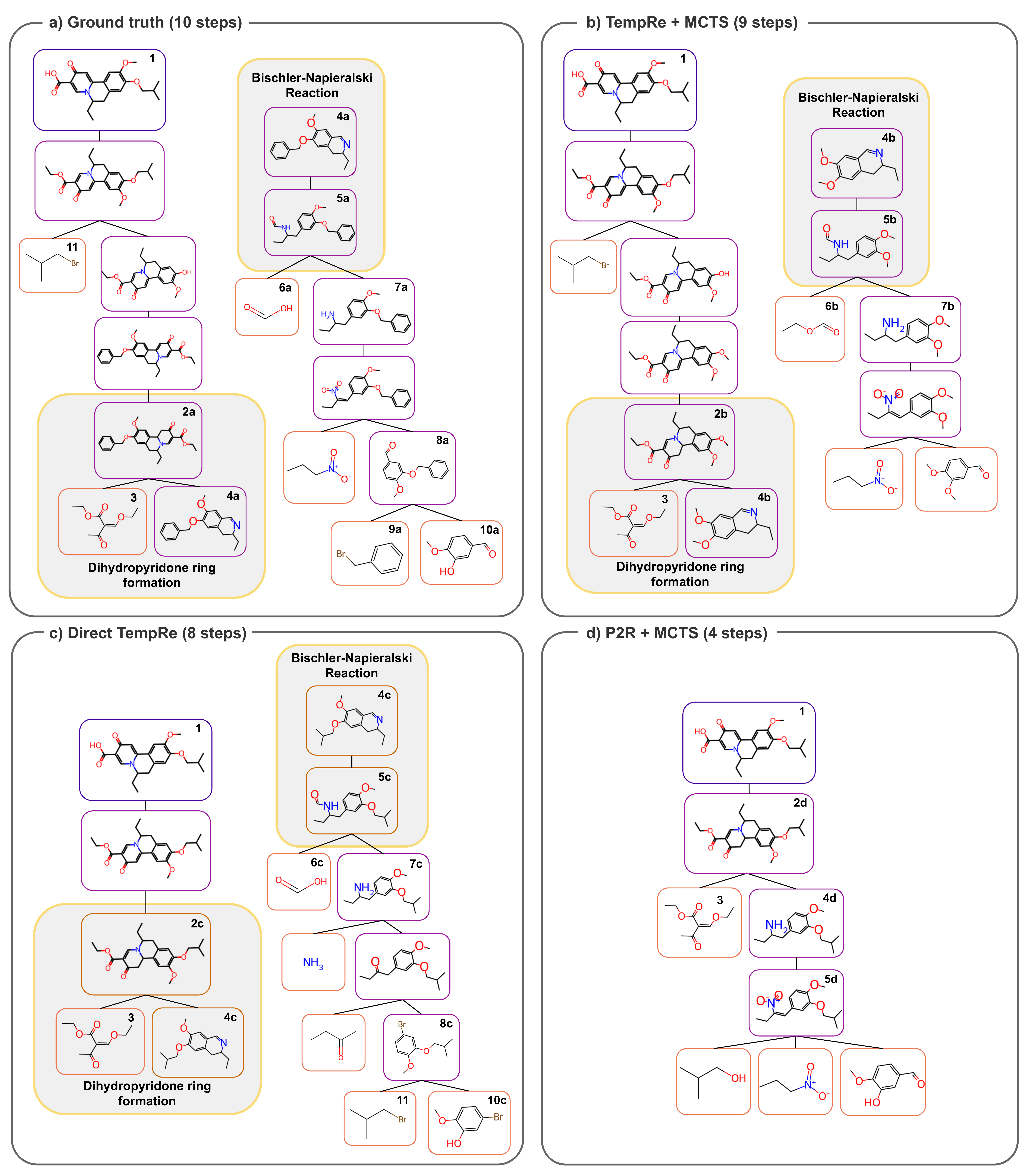}
    \caption{(a) Groundtruth synthetic route for dihydrobenzo[a]quinolizine \textbf{1}, and rank 1 route predictions of (b) P2T-Tok-Strict, (c) Direct TempRe, and (d) P2R.}
    \label{fig:case_study}
\end{figure}

To probe the behaviour of the models on complex targets, we undertook a case study on a 10-step synthesis from the n5 test set: the quinolizine derivative~\textbf{1}. This target presents a notable challenge, as its tricyclic fused-ring structure requires specific ring-forming reactions. The ground-truth route involves two main steps: a Bischler-Napieralski cyclisation to form a 3,4-dihydroisoquinoline intermediate (\textbf{4a}), which then reacts with an oxobutanoate (\textbf{3}) to complete the core scaffold (\textbf{2a}).

Whilst several models marked this target as `solved' (that is, they found a route to in-stock materials), none recovered the ground-truth path in their top-10 predictions. The predicted routes, however, reveal substantive differences in model behaviour (Figure~\ref{fig:case_study}).

The top-ranked prediction from P2T-Tok-Strict proposed a valid, in-stock synthesis in only nine steps. Notably, it employed the same core ring-forming strategy as the ground-truth route, including a Bischler-Napieralski cyclisation to yield the key intermediate~\textbf{4b}, which subsequently reacts with oxobutanoate~\textbf{3} to form the tricyclic product~\textbf{2b}.

Our Direct TempRe model also followed this successful strategy, yet proposed a more efficient, 8-step synthesis. The reduction in step-count was achieved by introducing the iso-butoxyl group early in the pathway. This group is stable and remains on the molecule throughout the synthesis, thereby avoiding the protection-deprotection sequence used in the ground-truth route (which involves protecting a hydroxyl on~\textbf{10a} with a benzyl group, before its eventual removal and replacement with the iso-butyl group). This result demonstrates TempRe’s ability not only to handle complex transformations but also to propose novel and chemically plausible routes that are more efficient than the recorded synthesis.

By contrast, the template-free P2R model suggested a 4-step route. Although this route reached in-stock starting materials, closer inspection revealed it to be chemically implausible. The proposed formation of the ring system~(\textbf{2d}) appears to be an over-simplification of the two distinct cyclisation reactions. Furthermore, the first step suggests an unlikely `two-in-one' transformation where alkylation and ether formation occur simultaneously to yield intermediate~\textbf{5d}. These observations highlight a notable limitation of such template-free models: a tendency to generate chemically unrealistic `shortcut' reactions that exploit the MCTS reward function. This can lead to an over-estimation of the true solve-rate and an under-estimation of realistic route lengths


\section{Conclusion}

In this work, we have introduced and evaluated TempRe, a framework that reformulates template-based retrosynthesis as a generative sequence modelling task. By generating reaction templates directly, TempRe establishes an effective middle ground between the scalability of template-free methods and the chemical robustness of traditional template classification.

Our comprehensive evaluations demonstrate the advantages of this approach. In single-step benchmarks, TempRe models consistently outperformed a template classifier, particularly on reactions involving rare templates. Crucially, in an out-of-distribution setting with large molecules, TempRe proved significantly more robust than a direct product-to-reactant model, mitigating a known generalisation failure of SMILES-based generation.

When applied to multi-step planning, a TempRe model constrained to known templates (P2T-Tok-Strict) achieved strong top-k route accuracy on the PaRoutes benchmark. This highlights a key finding: whilst unrestricted generation maximises the solve rate, constraining the output to a known chemical space yields higher-quality, more reliable synthetic routes. Furthermore, we extended the framework to direct multi-step synthesis, showing that representing entire routes as a sequence of templates is a promising and lightweight alternative to search. The case study analysis underscored the practical benefit of this, as TempRe models proposed chemically plausible and even novel, efficient routes, in contrast to the chemically questionable `shortcuts' generated by the template-free model.

Promising avenues for future work include the integration of reward mechanisms into the direct generation framework and the application of TempRe to other sequence architectures, such as large language models. The robustness demonstrated here suggests the framework is well-suited for broader application, including planning with noisier or more diverse reaction datasets beyond the USPTO corpus.

In summary, TempRe represents a significant step forward in computer-aided synthesis planning. By demonstrating the power of generative template modelling, it paves the way for the development of more generalisable, efficient, and chemically intuitive retrosynthesis tools.

\section{Acknowledgement}

We thank the Swiss National Science Foundation (SNSF) for funding this work [214915]. In addition we thank the  ChEMinformatics+ program for support in the project.

\clearpage
\bibliography{ref}

\clearpage
\appendix
\counterwithin{figure}{section}
\counterwithin{table}{section}

\section{Appendix}
\subsection{Experimental Setup and Implementation Details}
\label{app:data_details}

\subsubsection{General Data Processing and Filtering}
Our data processing pipeline, based on ReactionUtils and AiZynthTrain, begins with the USPTO dataset \cite{lowe2012extraction}. We performed atom-to-atom mapping using RXNMapper \cite{schwaller2021extraction}, removed spectator reagents, and extracted reaction templates with RDChiral \cite{coley2019rdchiral}. 
Unlike common procedures that discard low-frequency templates, we retained all reactions and implemented a set of filters to remove noisy or faulty ones. We keep reactions that satisfy the following criteria, or we modify the reaction if needed:
\begin{itemize}
    \item Number of reactants is no more than 3.
    \item Number of products is 1.
    \item Number of reactants' atoms is in the range [10, 70].
    \item Number of products' atoms is no less than 8.
    \item Total number of atoms in reactants is less than 4 times that in products.
    \item Number of unmapped atoms in reactants is less than 30.
    \item Remove reactants that do not contribute atoms to the products based on the atom mappings.
    \item Number of orphan atoms in the reactants or products is no more than 1. Orphan atoms are atoms with atom mapping on one side but not the other of a reaction.
    \item Number of unmapped atoms in the maximum common substructure (MCS) between the reactants and products is no more than 10.
    \item Product is not found in the reactants.
    \item No aromatic bonds between a mapped and an unmapped atom.
\end{itemize}
These filters remove only 9K reactions out of 1,058K, resulting in a final processed set of 1,050K reactions with 244K unique templates. As shown in Table \ref{tab:filter_ablation}, a model trained with this filtered data has a slight boost in top-k accuracy across all k values on the PaRoutes test set.
\begin{table}[h!]
    \centering
    \caption{Ablation study of data filter on single-step top-k accuracy of P2T on the PaRoutes test set.}
    \label{tab:filter_ablation}
    \begin{tabular}{@{}clllll@{}}
    \toprule
    \multicolumn{1}{l}{Train data filter} & Top-1 & Top-3 & Top-5 & Top-10 & Top-20 \\ \midrule
    \cmark & 0.564 & 0.797 & 0.859 & 0.910 & 0.933 \\
    \xmark & 0.546 & 0.783 & 0.850 & 0.905 & 0.928 \\ \bottomrule
    \end{tabular}
\end{table}

\subsubsection{Single-Step Reaction Benchmarks}
\label{app:single_step_benchmarks}
\begin{figure}[h!]
    \centering
    \includegraphics[width=\textwidth]{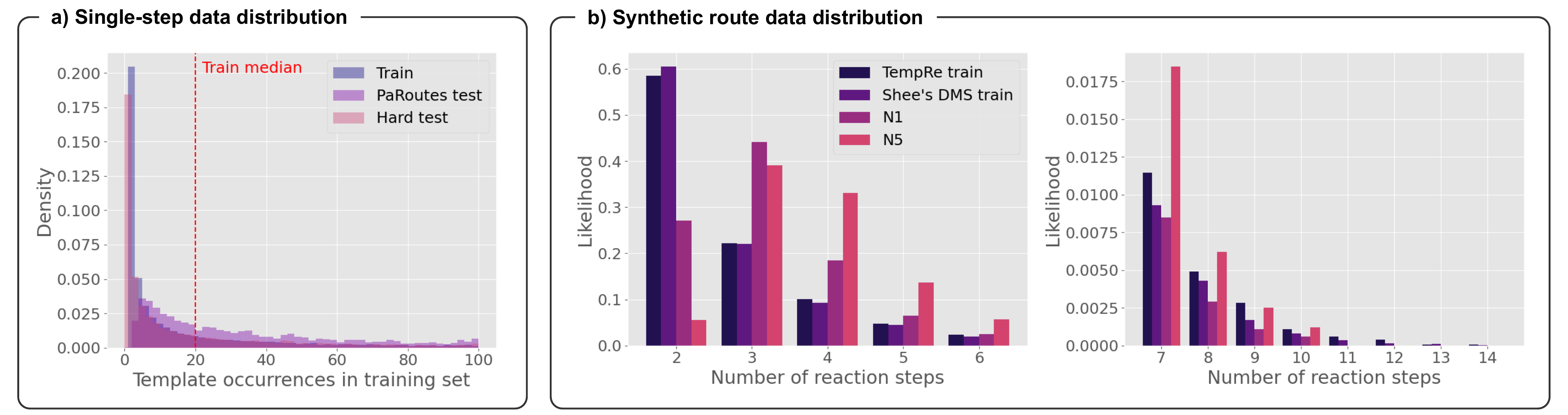}
    \caption{
        \textbf{Data distribution.} a) Distribution of training occurrences of templates in the training data, single-step PaRoutes test set and hard test set. For clarity, the distributions are clipped to 100 occurrences. b) Distribution of route lengths in the TempRe training set, DMS training set, test set n1, and test set n5.
    }
    \label{fig:data_distribution}
\end{figure}

From the filtered data, we constructed our final training and test sets. For testing, we used the PaRoutes benchmark \cite{genheden2022paroutes}, which consists of two non-exclusive test sets, N1 and N5. We extracted 55K single-step reactions from these routes for evaluating single-step models and ensured they were removed from our training data. To introduce a more challenging scenario, we also created a "hard test set" of 52K reactions from the remaining USPTO data, characterised by a higher proportion of rare templates compared to the PaRoutes set (Figure \ref{fig:data_distribution}a). Our final single-step dataset comprised a training set of 942K unique reactions (235K unique templates), the PaRoutes test set (55K reactions), and the hard test set (52K reactions).

\subsubsection{Out-of-Distribution (OOD) Benchmark}
To assess generalizability to OOD target molecules, we created an additional split of our filtered USPTO-Full data based on molecular weights. The training set consists of 800K reactions with an average product MW of $310 \pm 84$ Da, while the OOD test set contains 80K reactions with a significantly higher average product MW of $553 \pm 35$ Da.

\subsubsection{Multi-Step Route Dataset for Direct TempRe}
\begin{figure}[h]
    \centering
    \includegraphics[width=0.8\textwidth]{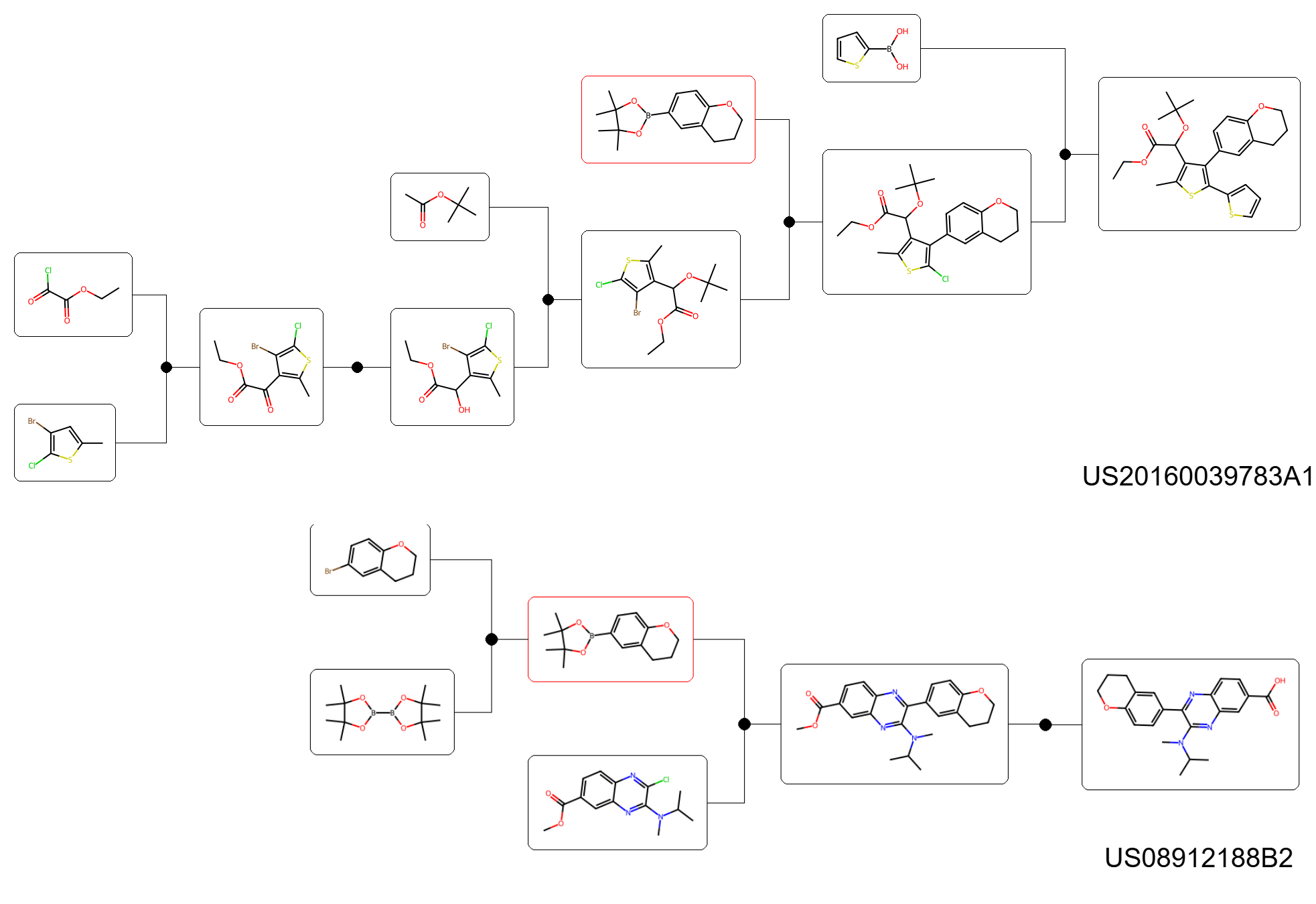}
    \caption{Example of a pair of overlapped routes. The red-highlighted leaf molecule of the first route is found as an intermediate in the second one. Such shorter routes were removed from the training set.}
    \label{fig:overlapped_routes_appendix}
\end{figure}

For training Direct TempRe models, we curated a dataset of 260K synthetic routes. This was done by grouping individual reactions from our processed USPTO data by patent ID and constructing synthetic trees via depth-first search. We then applied several filtering steps: we removed all single-step routes, routes containing looped reactions, and deduplicated routes. We also removed routes that were sub-routes of others to encourage termination at simpler, more likely starting materials (Figure \ref{fig:overlapped_routes_appendix}).
For comparison with an existing direct multi-step model \cite{shee2025directmultistep}, we also trained a Direct TempRe variant on their 164K synthetic route training set. The distribution of reaction steps across our training set, the DMS training set, and the PaRoutes test sets is shown in Figure \ref{fig:data_distribution}b, highlighting the strong bias towards shorter routes in all training data compared to the test benchmarks.


\subsection{Model Architectures and Training}
\subsubsection{Training and Model Details}
\label{app:training_details}
Our sequence-to-sequence models (P2T, P2T-Tok, P2R, and Direct TempRe variants) were implemented using the OpenNMT framework \cite{klein2017opennmt} with the Transformer architecture \cite{vaswani2017attention}. Each model was configured with 4 layers for both the Transformer encoder and decoder, 8 attention heads, an embedding size of 384, and a feed-forward layer size of 2048.

The models were trained using the Adam optimiser \cite{kingma2014adam} with  $\beta_1 = 0.9$ and $\beta_2 = 0.998$. A Noam decay learning rate schedule \cite{vaswani2017attention} was employed, featuring 8000 warm-up steps and a maximum learning rate of 2. To simulate larger effective batch sizes, gradient accumulation was used with an accumulation count of 4. Regularization was applied through dropout with a rate of 0.1, and model parameters were initialised using Glorot initialization \cite{glorot2010understanding}. For reproducibility, all models were trained with a fixed random seed (42).

Training was conducted for 200,000 steps with a batch size of 6144 tokens, which corresponds to approximately 8 epochs over our primary training dataset. All training and inference operations were performed on a single Nvidia GeForce RTX 3090 GPU.

For inference, beam search was employed. For single-step retrosynthesis predictions, a beam size of 100 was used. For multi-step planning (used for policy suggestions in MCTS and for Direct TempRe generation), a reduced beam size of 15 was applied to accelerate the search process.

\subsubsection{Tokenization Details}
\label{app:tokenisation}
For tokenizing template SMARTS, the P2T model used Byte Pair Encoding (BPE), resulting in a vocabulary size of 348. The P2T-Tok model employed a frequency-sensitive tokeniser: templates with more than 40 occurrences in the training set were encoded as single tokens. This compressed 235K unique reaction templates, with 401K reactions (43\% of the 942K training reactions) having their templates encoded as single tokens. BPE was then applied to the remaining templates, resulting in a total vocabulary size of 2651 tokens. Product SMILES for all sequence-to-sequence models were tokenised using a regular expression-based approach, resulting in a vocabulary of 121 tokens.

\subsection{Baseline Model Details}
\label{app:baseline_details}
\subsubsection{Baseline Implementations}
\begin{itemize}
\item \textbf{AiZynthFinder (AZF):} The AZF baseline \cite{genheden2020aizynthfinder} is a template-based retrosynthesis predictor that uses a Multi-Layer Perceptron (MLP) to classify which reaction template to apply. For our comparison, the model was required to parameterise our entire 235K template library.
\item \textbf{Product-to-Reactant (P2R) Transformer:} Our P2R baseline was implemented and trained with the same architecture and hyperparameters as our other generative models, as detailed above.
\item \textbf{Search-based Planners:} For multi-step planning, we used \textbf{MCTS}. In these frameworks, the AZF and P2R models function as the policy network to propose retrosynthetic disconnections at each step of the search.
\item \textbf{Direct Multi-Step (DMS):} This baseline \cite{shee2025directmultistep} is a generative transformer model for direct, end-to-end multi-step synthesis prediction.
\citet{shee2025directmultistep} introduced several variants that utilize different amounts of input information in addition to the target molecule. For a fair comparison, we opted to benchmark against the Explorer variant, which receives only the target molecule as input during inference.
\end{itemize}

\subsubsection{AiZynthFinder (AZF) Baseline Details}
\label{app:aizynth}
The AiZynthFinder (AZF) MLP classifier used as a baseline was trained on the same data. It takes a 1024-bit Morgan fingerprint \cite{morgan1965generation} of the target molecule as input. This input is mapped to a hidden vector of size 512 via a linear layer, and then to a probability distribution over the template library via a second linear layer. Given our training data's large template library of 235,000 unique templates, the output layer of the AZF model resulted in approximately 122 million parameters, highlighting the parameter inefficiency of classification models for such large template spaces.

\subsection{Search Algorithm Parameters}
\label{app:search_details}

We utilised the Monte Carlo Tree Search (MCTS) algorithm implemented in Syntheseus \cite{maziarz2025re} for search-based multi-step retrosynthesis. The selection phase of MCTS is guided by the Policy - Upper Confidence Bound (P-UCB) value:
\[
    \text{P-UCB}(s) = Q(s) + C_{\text{pucb}} \times \pi(s) \times \frac{\sqrt{N(p)}}{1+N(s)}
\]
where $Q(s)$ is the expected reward of the node, $C_{\text{pucb}}$ is an exploration constant, $\pi(s)$ is the prior probability of the node, $N(p)$ is the number of times the parent node $p$ has been visited, and $N(s)$ is the number of times the node $s$ has been visited.

At the beginning of the search, $Q(s)$ is initialised to 0.5 for all nodes. During the backpropagation phase, $Q(s)$ is updated every time a node is visited:
\[
    Q(s) \leftarrow \frac{r(s) + Q(s) \times N(s)}{N(s) + 1}
\]
where $r(s)$ is a reward function that returns 1 if at least one node in the subtree of $s$ contains a solution (i.e., all precursors are in the stock set), and 0 otherwise.

For our experiments, the exploration constant $C_{\text{pucb}}$ was set to 100. The policy $\pi(s)$ is derived from the logit normalization over the suggestions provided by the single-step policy model:
\[ 
    \pi(s) = \frac{\exp(\text{log }p_{\theta}(s)/T)}{\sum_{s' \in S} \exp(\text{log }p_{\theta}(s')/T)}
\]
where $S$ is the list of suggestions, and $p_{\theta}(s)$ is the probability of node $s$ assigned by the single-step model. The temperature $T$ was set to 3.0.

For each call to the single-step policy model during the expansion phase, we limited the number of suggestions (expansions $|S|$) to 10.

\paragraph{Hardware for Search Execution}
For each target molecule, the MCTS search was run with a limitation of 500 iterations or 300 seconds. The search algorithm itself was executed on $12 \times 2$ cores of AMD Ryzen 9 5900X processors, while calls to the single-step policy models were handled by the Nvidia GeForce RTX 3090 GPU mentioned in Appendix \ref{app:training_details}.

\subsection{Direct TempRe Algorithmic Details}
\label{app:algo}

\begin{algorithm}[H]
    \caption{Construct Molecular Set Graph from a Sequence of Reaction Templates}
    \label{alg:direct_tempre}
    \begin{algorithmic}[1]
    \REQUIRE Product molecule $p$ (as SMILES string), list of templates $\mathcal{T}$, optional stock inventory $\mathcal{S}$
    \ENSURE Molecular set graph $\mathcal{G}$
    
    \STATE Initialise root node $n_0 \leftarrow$ \texttt{MolSetNode} containing molecule $p$
    \IF{$\mathcal{S}$ is provided}
        \STATE Mark $p$ as purchasable if $p \in \mathcal{S}$
    \ENDIF
    \STATE Initialise graph $\mathcal{G}$ with root node $n_0$
    \STATE Initialise current node list $\mathcal{C} \leftarrow [n_0]$
    
    \FOR{each template $t \in \mathcal{T}$ with index $i$}
        \STATE Initialise next node list $\mathcal{N} \leftarrow [\,]$
        \FOR{each node $n \in \mathcal{C}$}
            \STATE $R \leftarrow$ \texttt{apply\_template\_to\_molecule\_set}($n.\text{mols}, t$)
            \IF{$R \neq \emptyset$}
                \STATE Expand $\mathcal{G}$ with $R$ from node $n$ (enforcing tree structure)
                \STATE Append resulting nodes to $\mathcal{N}$
            \ENDIF
        \ENDFOR
        \FOR{each node $n' \in \mathcal{N}$}
            \STATE Annotate $n'$ with template $t$ and index $i$
            \IF{$\mathcal{S}$ is provided}
                \STATE Mark each molecule in $n'.\text{mols}$ as purchasable if in $\mathcal{S}$
            \ENDIF
        \ENDFOR
        \IF{$\mathcal{N} \neq \emptyset$}
            \STATE Update $\mathcal{C} \leftarrow \mathcal{N}$
        \ENDIF
    \ENDFOR
    
    \RETURN $\mathcal{G}$
\end{algorithmic}
\end{algorithm}

\subsection{Single-step Retrosynthetic Analysis}





\subsubsection{Generation of Novel Templates}

\label{app:novel}

\begin{figure}[h!]
    \centering
    \includegraphics[width=\textwidth]{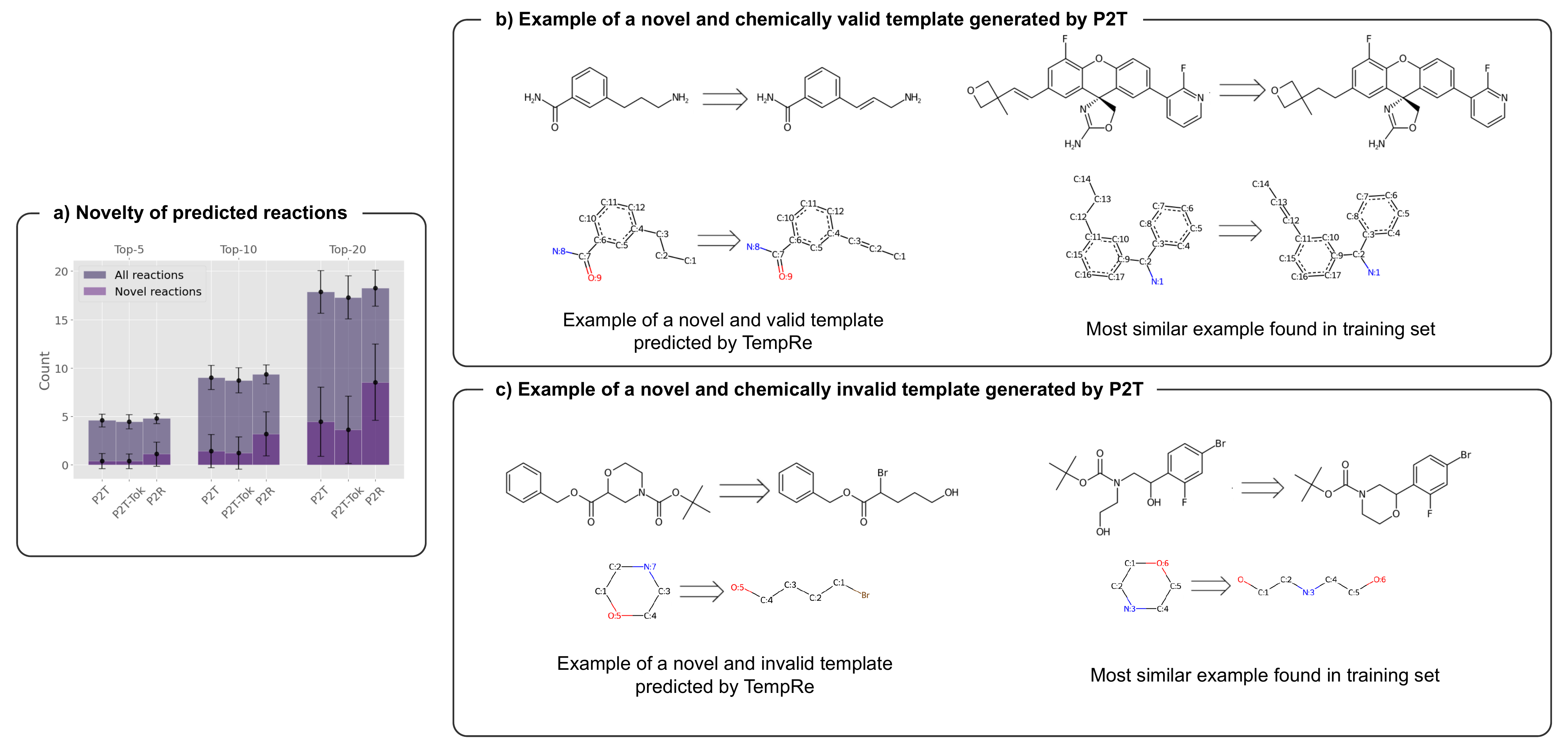}
    \caption{
        \textbf{Novelty of predicted reactions.} A reaction is considered novel if its template hash string is not found among those in the training data. a) Number of novel predicted reaction templates compared with the total number of predictions on the PaRoutes test set. The numbers of predictions are derived after deduplication and removing syntactically invalid templates. For P2R, atom-to-atom mapping is done by RXNMapper \cite{schwaller2021extraction}, followed by template extraction by RDChiral. The error bars represent standard deviation. b) Examples of chemically valid and invalid novel templates generated by P2T.
        }
    \label{fig:novel_reaction}
\end{figure}

Whilst generative in nature, the P2T and P2T-Tok models were observed to produce novel reaction templates not present in the training set. However, the frequency of such occurrences was found to be two to three times lower than for the P2R model (Figure~\ref{fig:novel_reaction}a). This suggests that the task of generating a reaction template imposes a greater degree of implicit constraint than the \emph{de-novo} generation of reactants. Consequently, post-processing filters can be more readily applied to TempRe outputs; for example, novel templates may be assessed for plausibility via a similarity search against the training library, or simply discarded, as in the Strict model variants. Applying an equivalent filtering process to P2R outputs is less straightforward, as it necessitates an atom-mapping step which introduces both computational overhead and potential ambiguity.

The models' capacity for generalisation is demonstrated in Figure~\ref{fig:novel_reaction}b, where P2T proposes a chemically valid dehydrogenation template. Although this exact template is novel, it is highly similar (Tanimoto~0.82) to a known template from the training set, indicating that the model has successfully adapted a learned reaction motif to a new chemical context. However, the generation of novel templates is not invariably sound. Figure~\ref{fig:novel_reaction}c illustrates an instance where P2T generates a chemically implausible template for a morpholine ring formation. This dichotomy between plausible generalisation and chemical invalidity underscores the utility of the library-constrained `Strict` variants, which allow the user to prioritise chemical fidelity over model creativity depending on the application.

\subsection{Search-Based Performance Details}

Here we provide a collection of Figures detailing additional information around our search-based performance.

\subsubsection{Multi-step solve time}

\begin{figure}[t]
    \centering
    \includegraphics[width=0.8\textwidth]{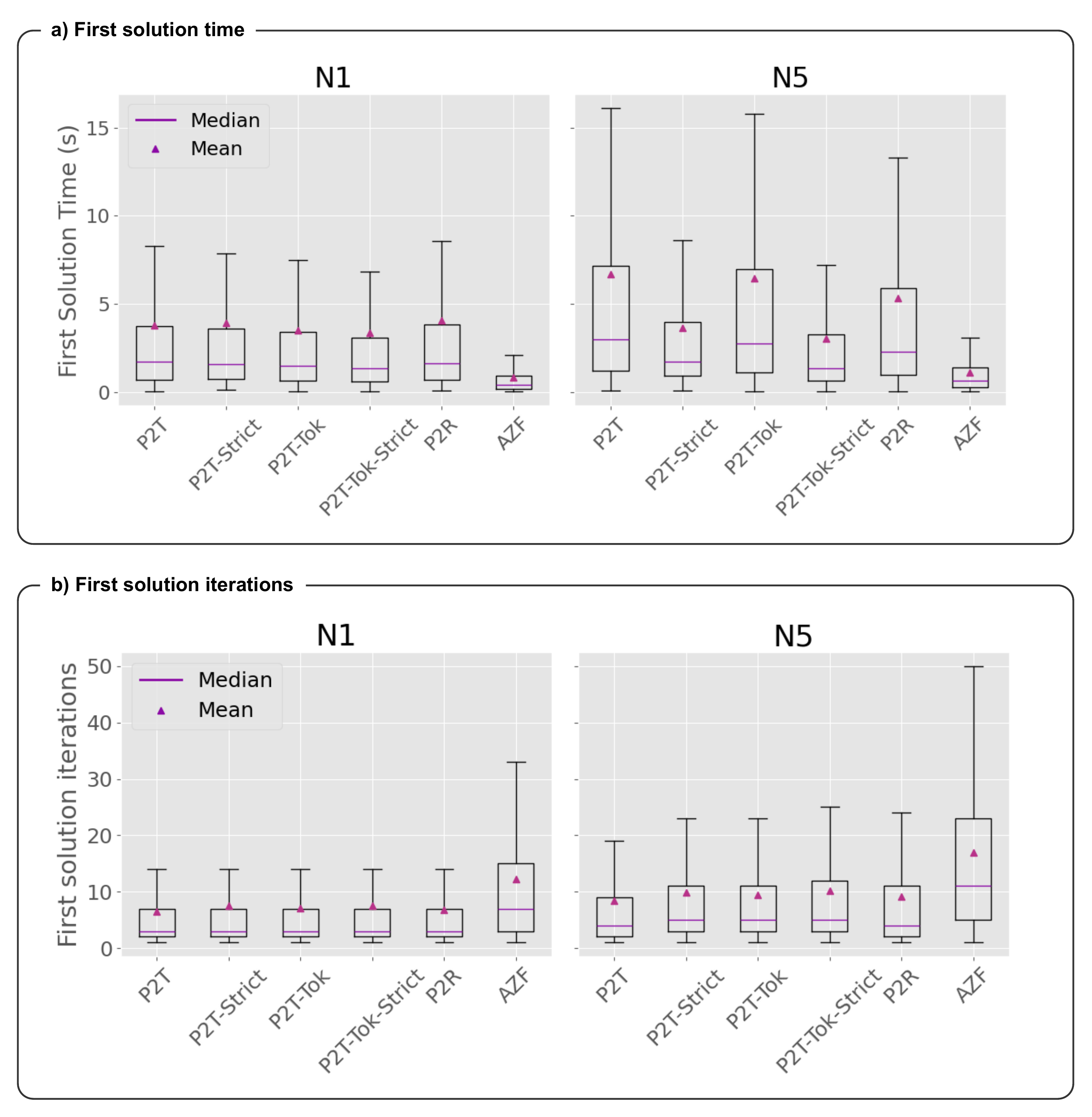}
    \caption{
        First solution time in seconds (a) and number of iterations (b). The boxes contain data between Q1 and Q3 quartiles, and the whiskers show minimum and maximum values without outliers.
    }
    \label{fig:first_solve_time_iter}
\end{figure}

First solution time and first solution number of iterations of all models are shown in Figure \ref{fig:first_solve_time_iter}.

\subsubsection{PaRoutes Stepwise Performance}

\begin{figure}[h]
    \centering
    \includegraphics[width=0.9\linewidth]{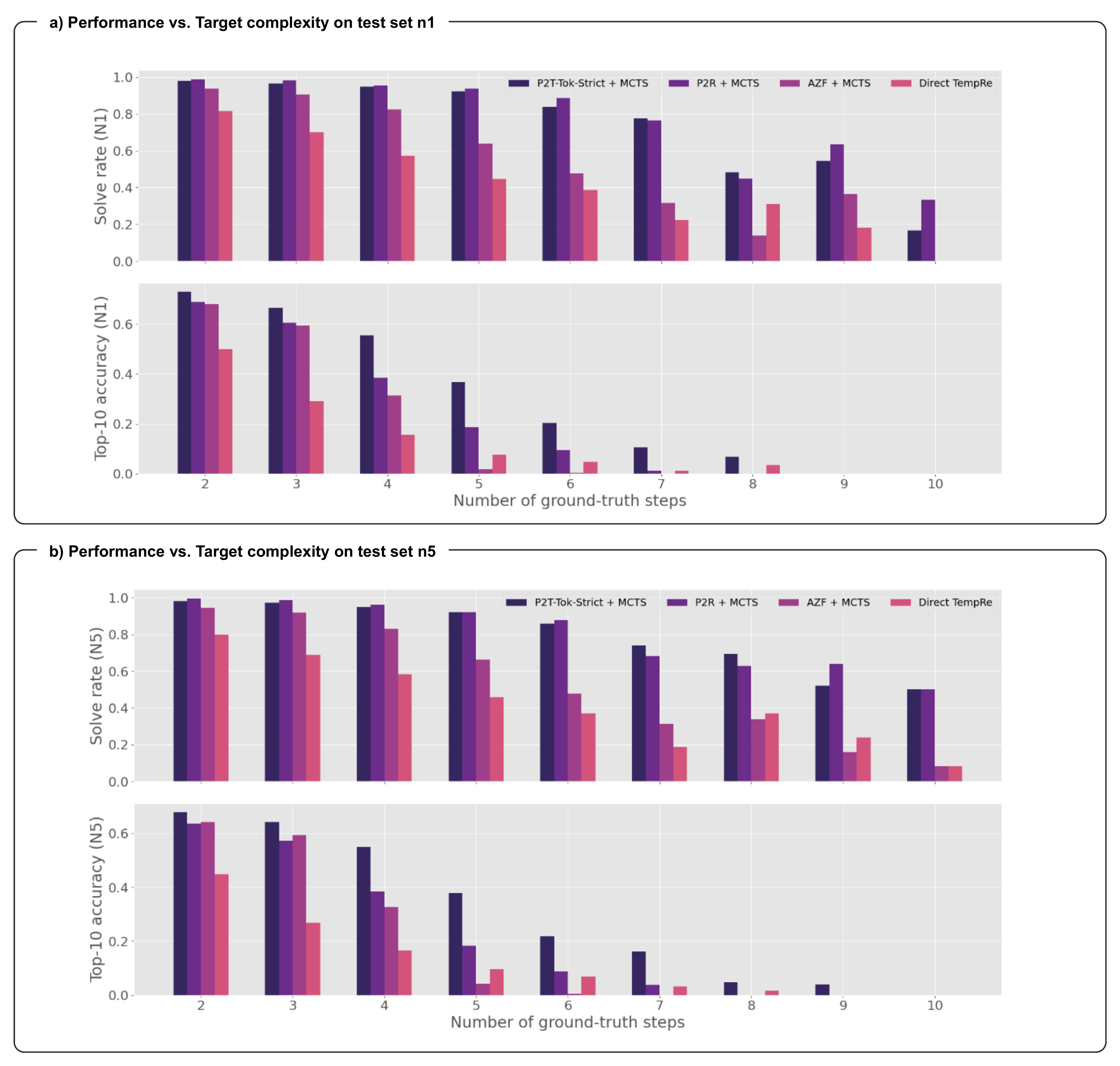}
    \\
    \centering
    \begin{tabular}{@{}lllllllllll@{}}
        \toprule
        \# reaction steps & 2 & 3 & 4 & 5 & 6 & 7 & 8 & 9 & 10 & Total \\ \midrule
        Set n1 size & 2709 & 4416 & 1842 & 647 & 255 & 85 & 29 & 11 & 6 & 10000 \\ 
        Set n5 size & 561 & 3907 & 3306 & 1372 & 570 & 185 & 62 & 25 & 12 & 10000 \\ \bottomrule
    \end{tabular}

    \caption{Distribution of solve-rate and top-10 route accuracy on sets (a) n1 and (b) n5 in correlation with the number of reaction steps of the ground-truth routes. The table shows the number of routes for each number of ground-truth steps.}
    \label{fig:stepwise_performance}
\end{figure}

Figures~\ref{fig:stepwise_performance} details the solve-rate and top-10 accuracy, correlated with the number of reaction steps in the ground-truth routes. A consistent trend is observed across all models and on both datasets (N1 and N5): model performance degrades as the length of the synthetic route increases. This suggests that longer, and therefore more complex, syntheses pose a greater challenge for all evaluated methods. The accompanying tables show that the datasets are dominated by routes with fewer than 6 steps, which is the region where the models perform best. This skew explains the high overall performance metrics, as the models are most frequently evaluated on the problems they are best at solving.

\label{app:Lengths}

\begin{figure}[h!]
    \centering
    \includegraphics[width=0.9\linewidth]{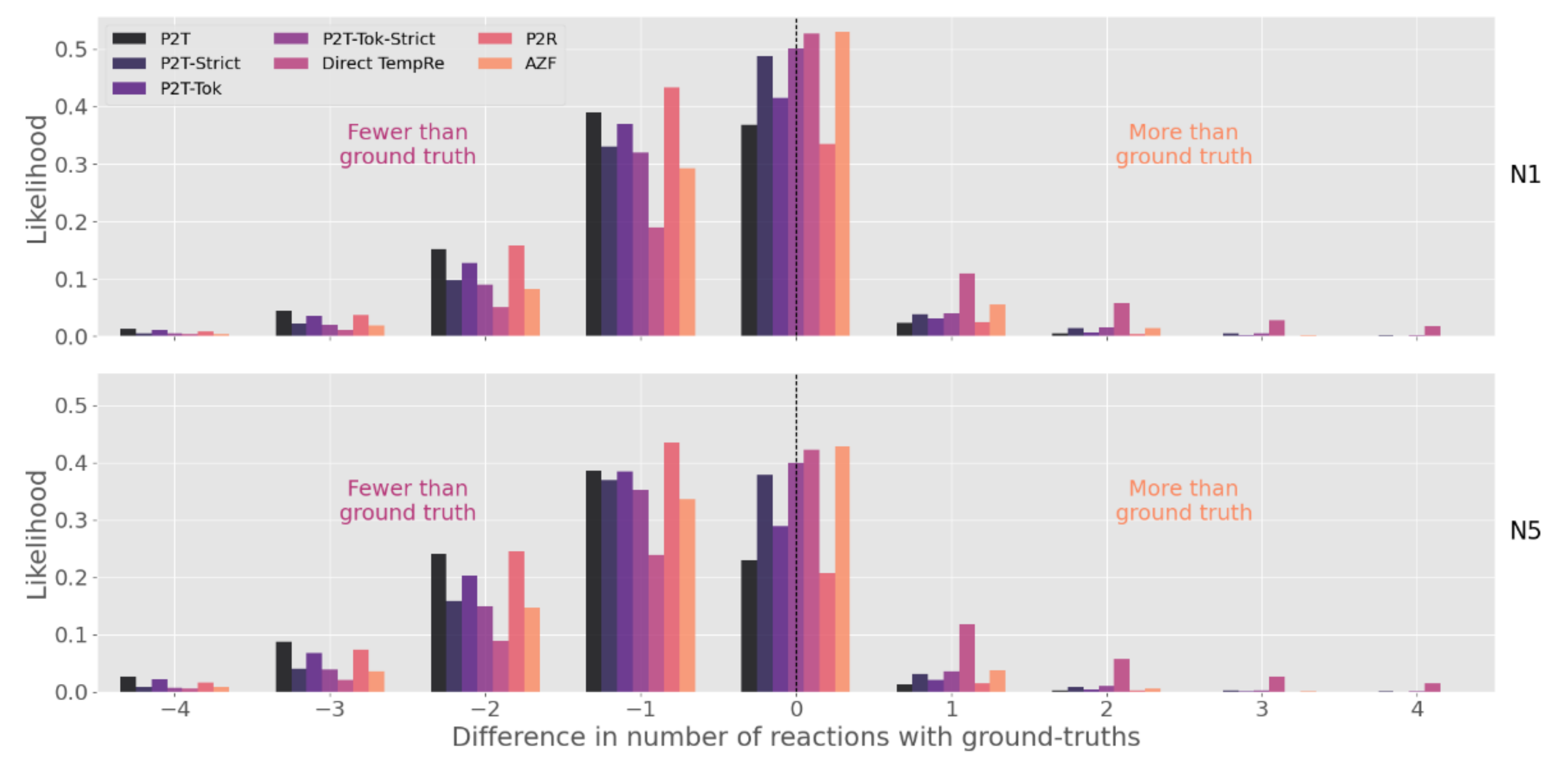}
    \caption{Difference in number of reactions between the rank 1 predicted routes and the corresponding ground-truth routes.}
    \label{fig:route_len}
\end{figure}



Figure~\ref{fig:route_len} illustrates the difference in the number of reactions between predicted and ground-truth routes. The problem setting is Model included in an iterative MCTS algorithm. In general, the models produce shorter routes than the ground truth, as indicated by the higher likelihoods for negative differences (fewer reactions). The extent to which the predicted routes are shorter varies among the different models, with AZF, P2T-Tok-Strict, and Direct TempRe producing the routes most similar in length to the ground truth.

\subsubsection{Effect of Expansion Size in MCTS}
\label{app:expansion_effect}

When combining single-step model with a search algorithm, it is common to get 50 candidates for each expansion step \cite{segler2018planning,genheden2020aizynthfinder,torren2024models}.
However, in our experiments, we chose the number of expansions to be only 10. 
For sequence-to-sequence models, having smaller number of expansions requires a smaller beam size for beam search, which could accelerate the inference speed.
Even for AiZynthFinder, we found that using 50 expansions decreases its performance on set n1 (Table \ref{tab:n_expansions}).

\begin{table}[h]
    \centering
    \caption{Comparison of solve-rate and top-k accuracy between 10 expansions and 50 expansions for MCTS with AiZynthFinder on test set n1}
    \label{tab:n_expansions}
    \begin{tabular}{@{}lllll@{}}
    \toprule
    \# expansions & Solve rate & Top-1 & Top-5 & Top-10 \\ \midrule
    10 & 0.86 & 0.24 & 0.49 & 0.51 \\
    50 & 0.81 & 0.17 & 0.38 & 0.40 \\ \bottomrule
    \end{tabular}
    \end{table}
\newpage
\subsection{Direct TempRe Variants and Inference Strategies}
\label{app:direct_tempre_variants}
We explored several variants of Direct TempRe to address the inherent short-route bias in multi-step synthesis training data. These variants differ in their conditioning during training and their inference strategies, aiming to encourage the generation of longer or more targeted synthetic routes. The strategies are summarized in Table~\ref{tab:direct_variants_appendix}.

\begin{table}[h]
    \centering
    \caption{Variants of Direct TempRe, their training conditions, and inference strategies.}
    \label{tab:direct_variants_appendix}
    \resizebox{\textwidth}{!}{%
    \begin{tabular}{@{}lllll@{}}
    \toprule
    \begin{tabular}[c]{@{}l@{}}Direct TempRe \\ variants\end{tabular} & \begin{tabular}[c]{@{}l@{}}Model\\ size\end{tabular} & \begin{tabular}[c]{@{}l@{}}Additional input \\ during training\end{tabular} & \begin{tabular}[c]{@{}l@{}}Inference\\ strategies\end{tabular} & \begin{tabular}[c]{@{}l@{}}Sampling\\ size\end{tabular} \\ \midrule
    Vanilla  & 20M & None & Sample 50 sequences & 50 \\
    N-step  & 20M & \# reaction steps & \begin{tabular}[c]{@{}l@{}}For each number of steps from 2 to 9, \\ sample 10 sequences\end{tabular} & 80 \\
    9-step  & 20M & \# reaction steps & \begin{tabular}[c]{@{}l@{}}Fix number of steps at 9, \\ sample 50 sequences\end{tabular} & 50 \\
    LeafSize  & 20M & \begin{tabular}[c]{@{}l@{}}\# atoms of the \\ largest leaf molecule\end{tabular} & \begin{tabular}[c]{@{}l@{}}For each number of atoms (10, 15, ..., 40),\\ sample 10 sequences\end{tabular} & 70 \\
    \bottomrule
    \end{tabular}%
    }
\end{table}

\begin{figure}[h!]
    \centering
    \includegraphics[width=0.9\textwidth]{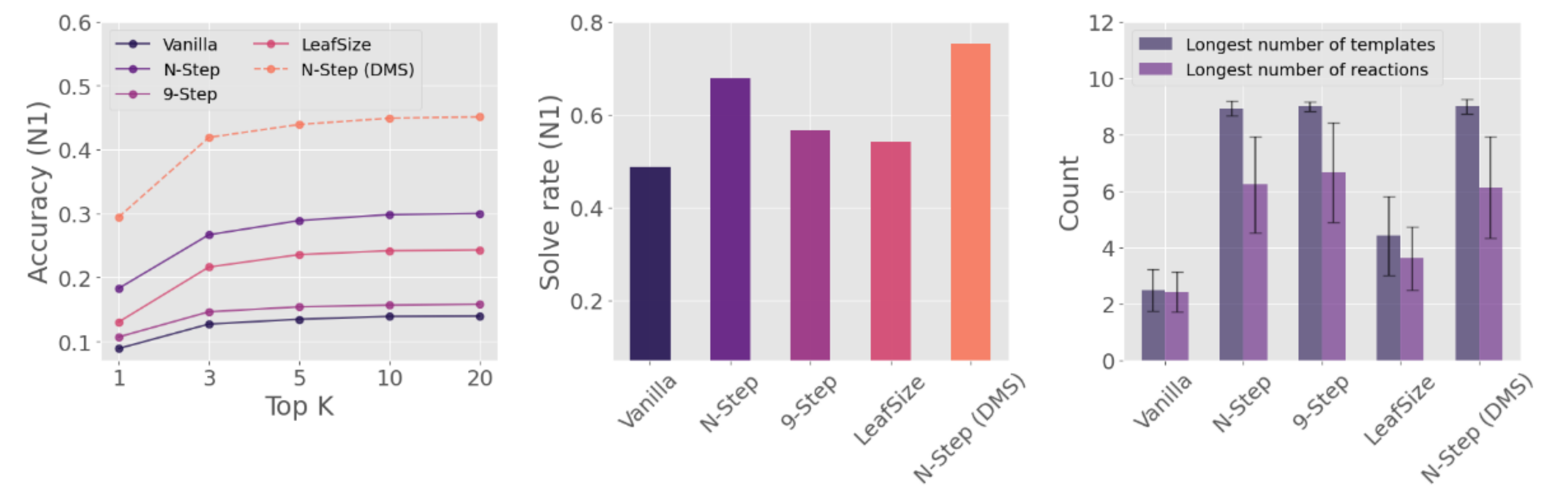}
    \caption{
        \textbf{Ablation study of Direct TempRe.} a) Top-k route accuracy and b) solve rate on test set n1. Note that N-Step (DMS) is plotted with a dashed line due to being trained on a different dataset than the others. c) Number of templates and the corresponding number of reactions of the longest predicted template sequence on set n1.
    }
    \label{fig:direct_tempre_ablation}
\end{figure}

We summarise the results of these models here and in  Figure \ref{fig:direct_tempre_ablation}
\begin{itemize}
    \item \textbf{Vanilla:} Standard training without any bias correction led to poor performance. The model tended to predict predominantly 2-step routes, reflecting the bias in the training data (Figure \ref{fig:direct_tempre_ablation}c) and resulting in a low solve rate.
    \item \textbf{Leaf / Size Conditioning:} Using the size (number of atoms) of the largest starting material (leaf molecule) as an additional input condition pushed the average number of generated templates to around 4 steps but showed limited improvement in overall performance.
    \item \textbf{N-step Conditioning:} Conditioning the model on the ground-truth number of steps during training and then calling the model multiple times during inference with different step counts (2-9) achieved the highest performance. This approach yielded a solve rate of 0.68 and a top-20 accuracy of 0.30 on our strictly processed data. The \textbf{9-step} variant, which only focused on long routes, did not perform as well, suggesting that the flexibility of scanning across multiple route lengths is crucial.
\end{itemize}
The superior performance of the N-step variant demonstrates the importance of explicitly addressing route length bias during both training and inference. To reconstruct synthetic routes from the generated template sequences, we iteratively apply each template to the current set of molecules. This process is detailed in Algorithm \ref{alg:direct_tempre}. The gap between the number of generated templates and actual decoded reactions occurs because sequence decoding stops when an invalid template is encountered.

We compared two different data processing strategies for training the best-performing N-step model.
\paragraph{TempRe Data Processing (Strict)}
Our approach removed all single-step reactions found in the PaRoutes test sets (n1 and n5) from the reaction pool before constructing training synthetic routes. This ensures no direct overlap between individual training reactions and test set reactions, providing a cleaner and more conservative evaluation of the model's ability to generalize to novel synthetic challenges. The N-step model trained on this data (Direct TempRe) achieved a 0.68 solve rate and 0.30 top-20 accuracy.

\paragraph{DMS Data Processing (Permissive)}
The approach used by \citet{shee2025directmultistep} splits data only at the route level, meaning some single-step reactions present in the test sets might also appear within training routes. While this represents a form of data leakage, it may not be necessarily detrimental, as the synthesis of common intermediates could be considered trivial knowledge. When trained on this data, our Direct TempRe (trained on DMS data) model achieved a higher 0.75 solve rate and 0.44 top-5 accuracy. This performance boost likely stems from the additional training signal provided by the overlapping reactions, which helps the model learn common synthetic transformations more effectively.

Both variants are included in our analysis to provide a complete picture of direct multi-step retrosynthesis performance under different data processing regimes.

\label{app:direct_varient_results}

\end{document}